%% file: main.tex
\documentclass[11pt]{article}
\usepackage{acl2016}
\usepackage{times}
\usepackage{url}
\usepackage{latexsym}
\usepackage{diagbox}
\usepackage{graphicx}
\usepackage{gnuplot-lua-tikz}
\usepackage{amsmath}
\usepackage{amsfonts}

\def\parcite#1{\cite{#1}} 
\def\perscite#1{\newcite{#1}} 

\def\samp#1{`\textit{#1}'}
\def\equo#1{``#1''}

\def\abs#1{\lvert#1\rvert}
\newcommand{\R}{\mathbb{R}}

\def\task#1{\textsc{#1}}
\def\mt{\task{mt}}
\def\lm{\task{lm}}
\def\tg{\task{tag}}
\def\lem{\task{lem}}

\aclfinalcopy 
\def\aclpaperid{90} 

\title{An Exploration of Word Embedding Initialization\\
in Deep-Learning Tasks}

\author{Tom Kocmi \and Ond{\v{r}}ej Bojar \\
  Charles University, \\
  Faculty of Mathematics and Physics \\
  Institute of Formal and Applied Linguistics \\
  surname\texttt{@ufal.mff.cuni.cz} \\}

\date{}

\begin{document}
\maketitle

\begin{abstract}
Word embeddings are the interface between the world of discrete units of 
text processing and the continuous, differentiable world of neural networks. In 
this work, we examine various random and pretrained initialization methods 
for embeddings used in deep networks and their effect on the performance on 
four NLP tasks with both recurrent and convolutional architectures.
We confirm that pretrained embeddings are a little better than random
initialization, especially considering the speed of learning. On the other
hand, we do not see any significant difference between various methods of random
initialization, as long as the variance is kept reasonably low. High-variance
initialization prevents the network to use the space of embeddings and forces it
to use other free parameters to accomplish the task. We support this hypothesis
by observing the performance in learning lexical relations and by the fact that
the network can learn to perform reasonably in its task even with fixed random
embeddings.
\end{abstract}

\section{Introduction}

Embeddings or lookup tables \parcite{bengio2003neural} are used for units of 
different granularity, from characters \cite{DBLP:journals/corr/LeeCH16} to 
subword units \cite{bpe,wu2016google} up to words. 
In this paper, we focus solely on word embeddings (embeddings attached to individual token types in the text). In their highly dimensional 
vector space, word embeddings are capable of representing many aspects of 
similarities between words: semantic relations or morphological properties 
\parcite{mikolov,Kocmi2016} in one language or cross-lingually 
\cite{Luong-etal:naacl15:bivec}.

Embeddings are trained \emph{for a task}. In other words, the vectors that 
embeddings assign to each word type are almost never provided manually but always 
discovered automatically in a neural network trained to carry out a particular 
task. The well known embeddings are those by \perscite{mikolov}, where the task 
is to predict the word from its neighboring words (CBOW) or the neighbors from 
the given word (Skip-gram). Trained on a huge corpus, these ``Word2Vec'' embeddings 
show an interesting correspondence between lexical relations and arithmetic 
operations in the vector space. The most famous example is the following: 
\def\wove#1{v(\textit{#1})}
\[
\wove{king} - \wove{man} + \wove{woman} \approx \wove{queen}
\]
In other words, adding the vectors associated with the words \samp{king} and
\samp{woman} while 
subtracting \samp{man} should be equal to the vector associated with the word
\samp{queen}. We can also say that the difference vectors
$\wove{king}-\wove{queen}$ and
$\wove{man}-\wove{woman}$ are almost identical and describe the gender relationship.

Word2Vec is not trained with a goal of proper representation of relationships, therefore the absolute accuracy scores around 50\% do not allow to rely on these relation predictions. Still, it is a rather interesting property 
observed empirically in the learned space. Another extensive study of embedding space has been conducted by \perscite{hill2017representational}.

Word2Vec embeddings as well as GloVe embeddings \parcite{pennington2014glove} became very popular
and they were tested in many tasks, also because for English they can be simply downloaded
as pretrained on huge corpora. Word2Vec was trained on 100 
billion words Google News dataset\footnote{See
\url{https://code.google.com/archive/p/word2vec/}.}
and GloVe embeddings were trained on 6 billion words from the Wikipedia. 
Sometimes, they are used as a fixed mapping for a better robustness of the system 
\cite{kenter2015short}, but they are more often used to seed the embeddings in a 
system and they are further trained in the particular end-to-end application 
\cite{collobert2011natural,lample2016neural}. 

In practice, random initialization of embeddings is still more common than using
pretrained embeddings and it should be noted that pretrained embeddings are
not always better than random initialization 
\parcite{DBLP:journals/corr/DhingraLSC17}.

%

We are not aware of any study of the effects of various random embeddings
initializations on the training performance.

In the first part of the paper, we explore various English word embeddings initializations in four 
tasks: neural machine translation (denoted \mt{} in the following for
short), language modeling 
(\lm{}), part-of-speech tagging (\tg{}) and lemmatization (\lem{}), covering
both common styles of neural architectures: the recurrent and
convolutional neural networks, RNN and CNN, resp. 


In the second part, we explore the obtained embeddings spaces in an attempt to
better understand the networks have learned about word relations.





\section{Embeddings initialization}
\label{sec:initialization}

Given a vocabulary $V$ of words, \emph{embeddings} represent each word as a
dense vector of size $d$ (as opposed to \equo{one-hot} representation where each
word would be represented as a sparse vector of size $\abs{V}$ with all zeros except
for one element indicating the given word).
Formally, embeddings are stored in a matrix $E \in \R^{\abs{V}\times d}$.

For a given word type $w \in V$, a row is selected from $E$. Thus, $E$ is often referred to as word 
lookup table. The size of embeddings $d$ is often set between 100 and 1000 
\cite{bahdanau:2014:corr,google2017attention,facebook2017convolutional}.

\subsection{Initialization methods}

Many different methods can be used to initialize the values in $E$ at the
beginning of neural network training. We distinguish between randomly
initialized and pretrained embeddings, where the latter
can be further divided into 
embeddings pretrained on the same task and pretrained on a standard task such as 
Word2Vec or GloVe.

Random initialization methods common in the literature\footnote{Aside from
related NN task papers such as \perscite{bahdanau:2014:corr} or
\perscite{facebook2017convolutional}, we also 
checked several popular neural network frameworks (TensorFlow, Theano, Torch, ...) to 
collect various initialization parameters.}
sample values either uniformly from a fixed interval centered at zero or, more often, from a zero-mean normal distribution with the standard deviation varying from 0.001 to 10.

The parameters of the distribution can be set empirically or calculated based on
some assumptions 
about the training of the network. The second approach has been done for various 
hidden layer initializations (i.e. not the embedding
layer). E.g. \perscite{xavier:init} and \perscite{he:init} argue that sustaining variance of values thorough the whole network leads to the best
results and  
define the parameters for initialization so that the layer weights 
$W$ have the same variance of output as is the variance of the input.

\perscite{xavier:init} define the \equo{Xavier} initialization method. They suppose a linear neural network 
for which they derive weights initialization as 
\begin{equation}
W \sim \mathcal{U} \big[ - \frac{\sqrt{6}}{\sqrt{n_i + n_o}}; \frac{\sqrt{6}}{\sqrt{n_i + n_o}}\big]
\end{equation}
where $n_i$ is the size of the input and $n_o$ is the size of the output. The initialization 
for nonlinear networks using ReLu units has been derived similarly by \perscite{he:init} as 
\begin{equation}
W \sim \mathcal{N}(0,\frac{2}{n_i})
\end{equation}
The same assumption of sustaining variance cannot be applied to embeddings
because there is no input signal whose variance should be sustained to the
output. We nevertheless try these initialization as well, denoting them
\emph{Xavier} and \emph{He}, respectively.


\subsection{Pretrained embeddings}

Pretrained embeddings, as opposed to random initialization, could work better,
because they already contain some information about word relations.

To obtain pretrained embeddings, we can train a randomly initialized model from the normal distribution with a standard deviation of 0.01, 
extract embeddings from the final model and use them as pretrained embeddings for the following 
trainings on the same task. Such embeddings contain information useful for the 
task in question and we refer to them as \emph{self-pretrain}.

A more common approach is to download some ready-made \equo{generic} embeddings
such as Word2Vec and GloVe, whose are not directly related to the final task but show to
contain many morpho-syntactic relations between words
\parcite{mikolov,Kocmi2016}.
%
%
Those embeddings are trained on billions of monolingual examples and can be
easily reused in most existing neural architectures.

\section{Experimental setup}

This section describes
the neural models we use for our four tasks
and the training and testing datasets.




\subsection{Models description}

For all our four tasks (\mt, \lm, \tg, and \lem), we use Neural Monkey \parcite{NeuralMonkey:2017}, an open-source neural machine
translation and general sequence-to-sequence learning system built using the
TensorFlow machine learning library.

All models use the same vocabulary of 50000 most frequent words from the training 
corpus. And the size of embedding is set to 300, to match the dimensionality of 
the available pretrained Word2Vec and GloVe embeddings.

All tasks are trained using 
the Adam \cite{journals-corr-KingmaB14} optimization algorithm.

We are using 4GB machine translation setup (\mt) as described in \perscite{cuni_training_wmt_2017} with increased encoder and decoder RNN sizes.
The setup is 
the encoder-decoder architecture 
with attention mechanism as proposed by \perscite{bahdanau:2014:corr}. We use encoder RNN with  500 GRU cells for each direction (forward and backward), decoder RNN with 450 conditional 
GRU cells, maximal length of 50 words and no dropout. We evaluate the performance using BLEU
\parcite{papineni:2002}.
Because our aim is not to surpass the state-of-the-art \mt{} performance, we omit
common extensions like beam search or ensembling. Pretrained embeddings
also prevent us from using subword units \parcite{bpe} or a larger embedding
size, as customary in NMT.
We experiment only with English-to-Czech \mt{} and when using pretrained embeddings we modify only the source-side
(encoder) embeddings, because there are no pretrained embeddings available for
Czech.

The goal of the language model (\lm) is to predict the next word based on
the history of previous words. Language modeling can be thus seen as (neural) machine
translation without the encoder part: no source sentence is given to translate
and we only 
predict words conditioned on the previous word and the state computed from predicted words.
Therefore the parameters of the neural network are the same as for the \mt{} decoder. 
The only difference is that we use dropout with keep probability of 0.7 
\cite{dropout}.
The generated sentence is evaluated as the perplexity against the gold output words (English in our case).


The third task is the POS tagging (\tg). We use our custom network architecture: The model starts with a bidirectional encoder 
as in \mt{}. For each encoder state, a fully connected linear layer then predicts a tag.
The parameters are set to be equal to 
the encoder in \mt{}, the predicting layer have a size equal to the number of tags.
\tg{} is evaluated by the accuracy of predicting the correct POS tag.

The last task examined in this paper is the lemmatization of words in a given
sentence (\lem).
For this task we have decided to use 
the convolutional neural network, which is second most used architecture in neural language processing next to the recurrent neural networks. We use the 
convolutional encoder as defined by \perscite{facebook2017convolutional} and for 
each state of the encoder, we predict the lemma with a fully connected linear layer. 
The parameters are identical to the cited work. \lem{} is evaluated by a accuracy of predicting the correct lemma.

When using pretrained Word2Vec and GloVe embeddings, we face the problem of different 
vocabularies not compatible with ours. Therefore for words in our vocabulary not covered by the pretrained
embeddings, we sample the embeddings
from the zero-mean normal distribution with a standard deviation of $0.01$.

\subsection{Training and testing datasets}

We use CzEng 1.6 \parcite{czeng16:2016}, a parallel Czech-English 
corpus containing over 62.5 million sentence pairs. This dataset 
already contains automatic word lemmas and POS tags.\footnote{We are
aware that training and evaluating a POS tagger and lemmatizer on automatically
annotated data is a little questionable because the data may exhibit artificial
regularities and cannot lead to the best performance, but we assume that this difference will have no effect on the
comparison of embeddings initializations and we prefer to use the same training dataset
for all our tasks.}

\begin{table*}[t]
\begin{center}
\small
\begin{tabular}{l|cccc}
Initialization & \mt{} en-cs (25M) & \lm{} (25M) & RNN \tg{} (3M) & CNN \lem{} (3M)\\
\hline
$\mathcal{N}(0,10)$   & 6.93 BLEU  & 76.95 & 85.2 \%  & 48.4 \% \\
$\mathcal{N}(0,1)$     & 9.81 BLEU  & 61.36 & 87.9 \% & 94.4 \% \\
$\mathcal{N}(0,0.1)$   & 11.77 BLEU & 56.61 & 90.7 \% & 95.7 \% \\
$\mathcal{N}(0,0.01)$  & 11.77 BLEU & 56.37 & \textbf{90.8 \%} & \textbf{95.9 \%}\\
$\mathcal{N}(0,0.001)$  & \textbf{11.88 BLEU} & \textbf{55.66} & 90.5 \% & \textbf{95.9 \%}\\
Zeros                  & 11.65 BLEU & 56.34 & 90.7 \%  & \textbf{95.9 \%}\\
Ones                   & 10.63 BLEU & 62.04 & 90.2 \% & 95.7 \% \\
He init.            & 11.74 BLEU & 56.40 & 90.7 \% & 95.7 \% \\
Xavier init.        & 11.67 BLEU & 55.95 & \textbf{90.8 \%} & \textbf{95.9 \%} \\
\hline
Word2Vec        & 12.37 BLEU  & \textbf{54.43} & 90.9 \% & 95.7 \% \\
Word2Vec on trainset    & 11.74 BLEU & 54.63 & 90.8 \% & 95.6 \%\\
GloVe           & 11.90 BLEU  & 55.56 & 90.6 \% & 95.5 \% \\
Self pretrain   & \textbf{12.61 BLEU} & 54.56 & \textbf{91.1 \%} & \textbf{95.9 \%}
\end{tabular}
\end{center}
\caption{Task performance with various embedding initializations.
Except for \lm{}, higher is better. The best results for random (upper part) and pretrained (lower part) embedding initializations are in bold.
}
\label{tab:initresults}
\end{table*}

We use the \texttt{newstest2016} dataset from WMT 2016\footnote{http://www.statmt.org/wmt16/translation-task.html} 
as the testset for \mt{}, \lm{} and \lem{}. The size of the testset is 2999
sentence pairs containing 57 thousands Czech and 67 thousands English running words.

For \tg{}, we use manually annotated English tags from 
PCEDT\footnote{https://ufal.mff.cuni.cz/pcedt2.0/en/index.html}
\parcite{pcedt20:lrec2012}. From this dataset, we 
drop all sentences containing the tag \equo{-NONE-} which is not part of 
the standard tags. This leads to the testset of 13060 sentences of 228k running
words.

\section{Experiments}

In this section, we experimentally evaluate embedding initialization methods across 
four different tasks and two architectures: the recurrent and convolutional neural networks.

The experiments are performed on the NVidia GeForce 1080 graphic card. Note that
each run of \mt{} takes a week of training, \lm{} takes a day and a half and
\tg{} and \lem{} need several hours each.
We run the training for one epoch and evaluate the performance
regularly throughout the training on the described test set.
For \mt{} and \lm{}, the epoch amounts to 25M sentence pairs
and for \tg{} and \lem{} to 3M sentences. 
The epoch size is set empirically so that the models already reach a stable level of performance
and further improvement does not increase the performance too much.

\mt{} and \lm{} exhibit performance fluctuation throughout the training.
Therefore, we average the results over five
consecutive evaluation scores
spread across 
500k training examples to avoid local fluctuations. This can be seen as a simple 
smoothing
method.\footnote{See, e.g. \url{http://www.itl.nist.gov/div898/handbook/pmc/section4/pmc42.htm} from  \perscite{natrella2010nist} justifying the use of the simple average, provided that the series has leveled off, which holds in our case.}


\subsection{Final performance}
\label{sec:influenceinitialization}

In this section, we compare various initialization methods based on the final
performance reached in the individual tasks. Intuitively, one
would expect the best performance with
self-pretrained embeddings, followed by Word2Vec and GloVe. The random embeddings 
should perform worse.

Table \ref{tab:initresults} shows the influence of the embedding 
initialization on various tasks and architectures.

The rows $ones$ and $zeros$ specify the initialization with a single fixed value.

The ``Word2Vec on trainset''
are pretrained embeddings which we created by running Gensim 
\cite{rehurek_lrec} on our training set. This setup serves as a baseline for the
embeddings pretrained on huge 
monolingual corpora and we can notice a small loss in performance compared to
both Word2Vec and GloVe.

We can notice several interesting results.
As expected, the self-pretrained embeddings 
slightly outperform pretrained Word2Vec and GloVe, which are generally slightly better than
random initializations.

A more important finding is that there is generally no significant 
difference in the performance between different random
initialization methods, except $ones$ and setups with the standard deviation of 1 and
higher, all of which perform considerably worse.

Any
random initialization with standard deviation smaller than 0.1 leads to 
similar results, including even the $zero$ initialization.\footnote{It 
could be seen as a surprise, that zero initialization works at all. But since 
embeddings behave as weights for bias values, they learn quickly from the random weights
available throughout the network.}
We attempt to explain this behavior in Section~\ref{sec:exploration_emb}. 

\subsection{Learning speed}

\begin{figure}
\begin{center}
\input{languagemodel.tex}
\end{center}
\caption{Learning curves for language modeling. The testing perplexity is computed every 300k training examples. Label "Remaining methods" represents all learning curves for the methods from Table \ref{tab:initresults} not mentioned otherwise.}
\label{fig:languagemodel}
\end{figure}
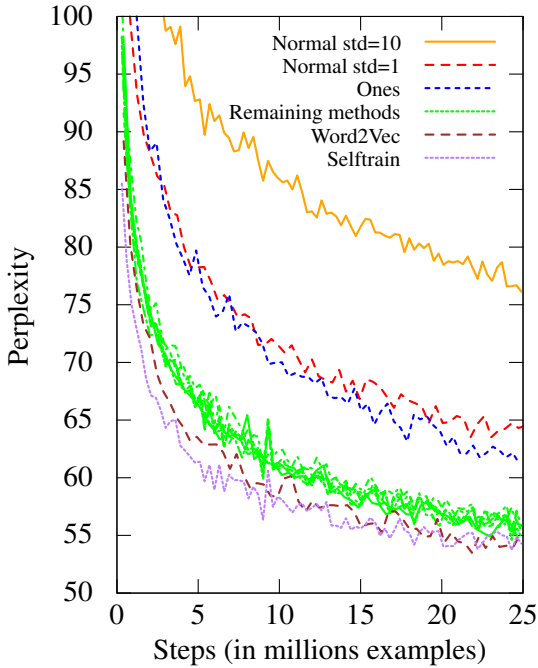

While we saw in Table~\ref{tab:initresults} that
most of the initialization methods lead to 
a similar performance, the course of the learning is slightly more varied. In
other words, different initializations need different numbers of training steps
to arrive at a particular performance. This is illustrated in
Figure~\ref{fig:languagemodel} for \lm{}.

To describe the situation concisely across the tasks, we
set a minimal 
score for each task and we measure how many examples did the training need to reach 
the score. We set the scores as follows: \mt{} needs to
reach 10 BLEU points, \lm{} needs to reach the perplexity of 60, \tg{} needs to reach
the accuracy of 90\% and \lem{} needs to reach the accuracy of 94\%.

We use a smoothing window as implemented in TensorBoard with a weight of 0.6 to 
smooth the testing results throughout the learning. This way, we avoid small fluctuations 
in training and our estimate when the desired value was reached is more reliable.

The results are in Table \ref{tab:rechingscore}. We can notice that pretrained embeddings 
converge faster than the randomly initialized ones on recurrent architecture
(\mt, \lm{} and \tg) but not on the convolutional architecture (\lem).

Self-pretrained embeddings are generally much better. Word2Vec also performs
very well but GloVe embeddings are worse than random initializations for \tg{}.

\begin{table}[t]
\begin{center}
\small
\begin{tabular}{l|cccc}
Initialization & \mt{} en-cs & \lm{} & \tg{} & \lem{}\\
\hline
$\mathcal{N}(0,1)$    & 25.3M & 37.3M & 10.6M & 2.7M \\
$\mathcal{N}(0,0.1)$   & 9.7M & 13.5M & 2.0M & 1.8M \\
$\mathcal{N}(0,0.01)$  & 9.8M & 12.0M & 1.4M & 1.2M \\
$\mathcal{N}(0,0.001)$ & 9.8M & 12.0M & 1.0M & 0.5M \\
Zeros                 & 9.4M & 12.3M & 1.0M & \textbf{0.5M} \\
Ones                  & 18.9M & 26.7M & 2.9M & 0.8M \\
He init.              & 9.5M & 12.5M & 1.0M & \textbf{0.5M} \\
Xavier init.          & 9.2M & 12.3M & 1.0M & \textbf{0.5M} \\
\hline
Word2Vec              & 6.9M & 7.9M & 0.7M & 1.2M \\
GloVe                 & 8.6M & 11.4M & 1.9M & 1.3M \\
Self pretrain         & \textbf{5.2M} & \textbf{5.7M} & \textbf{0.2M} & 0.9M
\end{tabular}
\end{center}
\caption{The number of training examples needed to reach a desired score.}
\label{tab:rechingscore}
\end{table}

\section{Exploration of embeddings space}
\label{sec:exploration_emb}

We saw above
that pretrained embeddings are slightly better than random 
initialization. We also saw that the differences in performance are not significant when
initialized randomly with
small values.

In this section, we analyze how specific lexical relations between words are
represented in the learned embeddings space. Moreover, based on the observations
from the previous section, we propose a hypothesis about the failure of
initialization with big numbers ($ones$ or high-variance random 
initialization) and try to justify it.

The hypothesis is as follows: 

The more variance the randomly initialized embeddings have, the more effort must
the
neural network exerts to store information in the embeddings space. Above a
certain effort threshold, it becomes easier to store the information 
in the subsequent hidden layers
(at the expense of some capacity loss) and use the random embeddings more or less as a
strange \equo{multi-hot} indexing mechanism. And on the other hand,
initialization 
with a small variance or even all zeros leaves the neural network free choice
over the utilization of the embedding space.


We support our hypothesis as follows. 
\begin{itemize}
\item We examine the embedding space on the performance in lexical relations
between words,
	If our hypothesis is plausible, low-variance
	embeddings will perform better at representing these relations.
\item We run an experiment with non-trainable fixed random initialization to demonstrate 
	the ability of the neural network to overcome broken embeddings and to learn the 
	information about words in its deeper hidden layers.
\end{itemize}

\subsection{Lexical relations}

Recent work on word embeddings \cite{vylomova,mikolov} has shown that simple vector operations 
over the embeddings are surprisingly effective at capturing various semantic and 
morphosyntactic relations, despite lacking explicit supervision in these
respects. 

The testset by \perscite{mikolov} contains \equo{questions} defined as $v(X)-v(Y)+v(A) \sim v(B)$. 
The well-known example involves predicting a vector for word \samp{queen} from the vector 
combination of $\wove{king}-\wove{man}+\wove{woman}$. This example is a part of
\equo{semantic relations} in the test set,
called opposite-gender. The dataset contain another 4 semantic relations and 9 morphosyntactic 
relations such as pluralisation  $\wove{cars}-\wove{car}+\wove{apple} \sim
\wove{apples}$.

\perscite{Kocmi2016} revealed the sparsity of the testset and presented 
extended testset. Both testsets are compatible and we use them in combination.

Note that the performance on this test set is affected by the vocabulary overlap
between the test set and the vocabulary of the embeddings; questions containing
out-of-vocabulary words cannot be evaluated. This is the main reason, why we
trained all 
tasks on the same training set and with the same vocabulary, so that their
performance in lexical relations can be directly compared.

Another lexical relation benchmark is the word similarity. The idea is that similar words such 
as \samp{football} and \samp{soccer} should have vectors close together. There exist many datasets 
dealing with word similarity. \perscite{faruqui-2014:SystemDemo} have extracted 
words similarity pairs from 12 
different corpora and created an interface for testing the embeddings on the word 
similarity task.\footnote{http://wordvectors.org/}

When evaluating the task, we calculate the similarity between a given pair
of words by the cosine similarity between their corresponding vector representation. We 
then report Spearman’s rank correlation coefficient between the rankings produced
by the embeddings against human rankings.
For convenience, we combine absolute values of Spearman's correlations from all
12 \perscite{faruqui-2014:SystemDemo} testsets
together as an average weighted by the number of words in the datasets.

The last type of relation we examine are the nearest neighbors. We illustrate on
the \tg{}
task how the embedding space is clustered when various initializations are used. We employ the Principal component analysis 
(PCA) to convert the embedding space of $\abs{E}$ dimensions into two-dimensional space.

\begin{table}[t]
\begin{center}
\small
\begin{tabular}{l|c|c|c}
Initialization & \mt{} en-cs & \lm{} & \lem{} \\
\hline
$\mathcal{N}(0,10)$    & 0.0; \phantom{1}0.3 & 0.0; \phantom{1}0.3 & 0.0; \phantom{1}0.3 \\
$\mathcal{N}(0,1)$     & 0.0; \phantom{1}0.4 & 1.4; \phantom{1}3.5 & 0.0; \phantom{1}0.3  \\
$\mathcal{N}(0,0.1)$   & 1.2; 23.5 & 5.5; 15.2 & 0.0; \phantom{1}0.8 \\
$\mathcal{N}(0,0.01)$  & 2.0; 29.9 & 6.9; 19.4 & 0.1; 32.7 \\
$\mathcal{N}(0,0.001)$ & 2.1; 31.4 & 6.7; 18.2 & 0.3; 33.3 \\
Zeros                  & 1.6; 29.5 & 6.0; 17.5 & 0.2; 31.1  \\
Ones                   & 0.5; 16.6 & 5.3; \phantom{1}9.3 & 0.1; 31.0 \\
He init.               & 1.4; 28.9 & 7.7; 18.3 & 0.1; 32.6 \\
Xavier init.           & 1.5; 29.5 & 7.4; 18.2 & 0.1; 32.7 \\
\hline
\hline
Word2Vec on trainset*  &\multicolumn{3}{c}{22.3; 48.9}  \\
Word2Vec official*     &\multicolumn{3}{c}{81.3; 70.7}  \\
GloVe official*        &\multicolumn{3}{c}{12.3; 60.1} \\
\end{tabular}
\end{center}
\caption{The accuracy in percent on the (semantic; morphosyntactic) questions. 
	We do not report \tg{} since its accuracy was less than 1\% on all questions.
	*For comparison, we present results of Word2Vec trained on our training set and official trained embeddings before applying them on training of particular task.}
\label{tab:questions}
\end{table}

\begin{table}[t]
\begin{center}
\small
\begin{tabular}{l|cccc}
Initialization & \mt{} en-cs & \lm{} & \tg{} & \lem{}\\
\hline
$\mathcal{N}(0,10)$    & 3.3 & 2.2 & 3.6 & 2.6\\
$\mathcal{N}(0,1)$     & 15.7 & 11.8 & 3.5 & 2.7\\
$\mathcal{N}(0,0.1)$   & 56.7 & 32.7 & 6.9 & 2.8\\
$\mathcal{N}(0,0.01)$  & 62.5 & 41.0 & 12.8 & 4.7\\
$\mathcal{N}(0,0.001)$ & 59.3 & 37.4 & 12.1 & 2.4\\
Zeros                  & 57.9 & 37.4 & 12.8 & 3.5\\
Ones                   & 34.0 & 19.3 & 11.4 & 4.3 \\
He init.               & 58.2 & 37.4 & 12.3 & 4.2 \\
Xavier init.           & 58.3 & 37.5 & 12.3 & 2.7
\end{tabular}
\end{center}
\caption{Spearman's correlation $\rho$ on word similarities. The results are multiplied by 100.}
\label{tab:wordsim}
\end{table}

\begin{figure*} 
  \includegraphics[width=\textwidth]{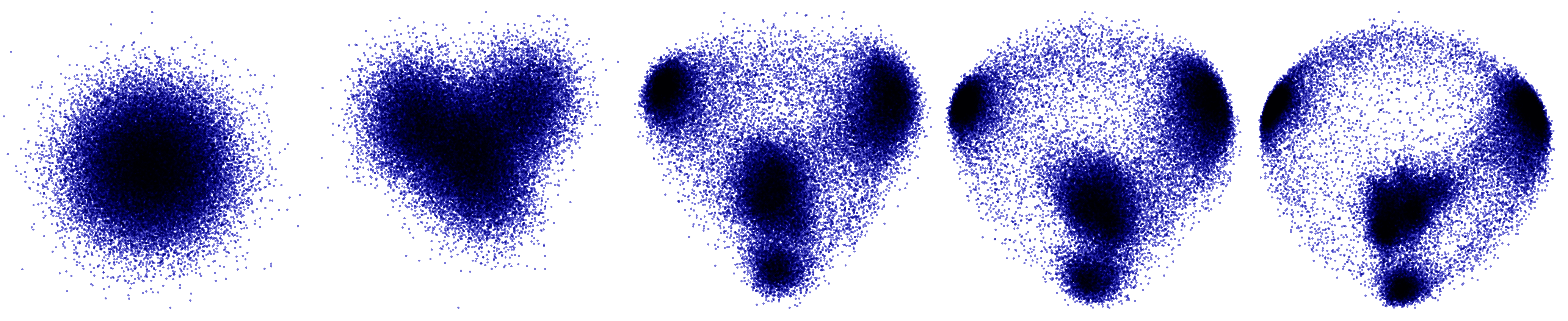}
\caption{
A representation of words in the trained embeddings for \tg{} task projected by PCA. From left to right it shows trained embeddings for 
$\mathcal{N}(0,1)$, $\mathcal{N}(0,0.1)$, $\mathcal{N}(0,0.01)$, $\mathcal{N}(0,0.001)$ 
and $zeros$. Note that except of the first model all of them reached a similar performance on the \tg{} task.}
\label{fig:nearestneighbors}
\end{figure*}

Table \ref{tab:questions} reflects several interesting properties about the embedding space. 
We see task-specific behavior, e.g. \tg{} not learning any of the tested relationships
whatsoever or \lm{} being the only task that learned at least something of
semantic relations.

The most interesting property is that when increasing the variance of initial embedding, 
the performance dramatically drops after some point.
\lem{} reveals this behavior the most: 
the network initialized by normal distribution with standard deviation of 
$0.1$ does not learn any relations but still performs comparably with other initialization 
methods as presented in Table~\ref{tab:initresults}. We ran the lemmatization experiments 
once again in order to confirm that it is not only a training fluctuation.

This behavior suggests that the neural network can work around broken embeddings and learn 
important features within other hidden layers instead of embeddings.

A similar behavior can be traced also in the word similarity evaluation in Table~\ref{tab:wordsim}, 
where models are able to learn to solve their tasks and still not learn any
information about word similarities in the embeddings.

Finally, when comparing the embedded space of embeddings as trained by \tg{} in
Figure~\ref{fig:nearestneighbors}, we see a similar behavior. With lower
variance in embeddings initialization, 
the learned embeddings are more clearly separated.

This suggests that when the neural network has enough freedom over the embeddings
space,
it uses it to store information about the relations between words.

\subsection{Non-trainable embeddings}

To conclude our hypothesis, we demonstrate the flexibility 
of a neural network to learn despite a broken embedding layer.

In this experiment, the embeddings are fixed 
and the neural network cannot modify them during the training process.
Therefore, it needs to find a way to learn the 
representation of words in other hidden layers.

As in Section~\ref{sec:influenceinitialization}, we train models for 3M examples 
for \tg{} and \lem{} and for over 25M examples for \mt{} and \lm{}.

\begin{table}[t]
\begin{center}
\small
\begin{tabular}{l|cccc}
Initialization & \mt{} en-cs & \lm{} & \tg{} & \lem{}\\
\hline
$\mathcal{N}(0,10)$    & 7.28 BLEU & 79.44 & 47.3 \% & 85.5 \%\\
$\mathcal{N}(0,1)$     & 8.46 BLEU & 78.68 & 87.1 \% & 94.0 \%\\
$\mathcal{N}(0,0.01)$  & 6.84 BLEU & 82.84 & 63.2 \% & 91.1 \%\\
Word2Vec               & 8.71 BLEU & 60.23 & 88.4 \% & 94.1 \%
\end{tabular}
\end{center}
\caption{The results of the experiment when learned with non-trainable embeddings.}
\label{tab:notrain}
\end{table}

Table \ref{tab:notrain} confirms that the neural network is flexible enough 
to partly overcome the problem with fixed embeddings. For example, \mt{} initialized with $\mathcal{N}(0,1)$ reaches the score of 8.46 BLEU with fixed embeddings compared to 9.81 BLEU for the same but not fixed (trainable) embeddings.

When embeddings are fixed at random values, the effect is very similar to
embeddings with high-variance random initialization.
The network can distinguish the words through the crippled embeddings but has no
way to improve them. It thus proceeds to learn 
in a similar fashion as with  one-hot representation.

\section{Conclusion}

In this paper, we compared several initialization methods of embeddings on four different tasks, namely: machine translation (RNN), language 
modeling (RNN), POS tagging (RNN) and lemmatization (CNN).

The experiments indicate that pretrained embeddings converge faster than random
initialization and 
that they reach a slightly better final performance.

The examined random initialization methods do not lead to significant
differences
in the performance as long as the initialization is within reasonable 
variance (i.e. standard deviation smaller than $0.1$). Higher variance
apparently prevents the network to adapt the embeddings to its needs and the network resorts
to learning in its other free parameters. We support this explanation by showing 
that the network is flexible enough to overcome even non-trainable embeddings.

We also showed a somewhat unintuitive result that when the neural 
network is presented with embeddings with a small variance or even all-zeros embeddings, 
it utilizes the space and learns (to some extent) relations between words in a
way similar to Word2Vec learning.


\section*{Acknowledgement}

This work has been in part supported by the European Union's Horizon 2020
research and innovation programme under grant agreements No 644402 (HimL) and 645452 (QT21),
by the LINDAT/CLARIN project of the Ministry of Education, Youth and Sports of
the Czech Republic (projects LM2015071 and OP VVV VI
CZ.02.1.01/0.0/0.0/16\_013/0001781), by the Charles University Research
Programme ``Progres'' Q18+Q48, by the Charles University SVV project number 260 453 and
by the grant GAUK 8502/2016.

\bibliography{biblio}
\bibliographystyle{acl2016}

\end{document}

%% file: languagemodel.tex
\begin{tikzpicture}[gnuplot]
\path (0.000,0.000) rectangle (7.400,4.500);
\gpcolor{color=gp lt color border}
\gpsetlinetype{gp lt border}
\gpsetlinewidth{1.00}
\draw[gp path] (1.504,0.985)--(1.684,0.985);
\draw[gp path] (6.847,0.985)--(6.667,0.985);
\node[gp node right] at (1.320,0.985) { 50};
\draw[gp path] (1.504,1.750)--(1.684,1.750);
\draw[gp path] (6.847,1.750)--(6.667,1.750);
\node[gp node right] at (1.320,1.750) { 55};
\draw[gp path] (1.504,2.514)--(1.684,2.514);
\draw[gp path] (6.847,2.514)--(6.667,2.514);
\node[gp node right] at (1.320,2.514) { 60};
\draw[gp path] (1.504,3.279)--(1.684,3.279);
\draw[gp path] (6.847,3.279)--(6.667,3.279);
\node[gp node right] at (1.320,3.279) { 65};
\draw[gp path] (1.504,4.043)--(1.684,4.043);
\draw[gp path] (6.847,4.043)--(6.667,4.043);
\node[gp node right] at (1.320,4.043) { 70};
\draw[gp path] (1.504,4.808)--(1.684,4.808);
\draw[gp path] (6.847,4.808)--(6.667,4.808);
\node[gp node right] at (1.320,4.808) { 75};
\draw[gp path] (1.504,5.572)--(1.684,5.572);
\draw[gp path] (6.847,5.572)--(6.667,5.572);
\node[gp node right] at (1.320,5.572) { 80};
\draw[gp path] (1.504,6.337)--(1.684,6.337);
\draw[gp path] (6.847,6.337)--(6.667,6.337);
\node[gp node right] at (1.320,6.337) { 85};
\draw[gp path] (1.504,7.101)--(1.684,7.101);
\draw[gp path] (6.847,7.101)--(6.667,7.101);
\node[gp node right] at (1.320,7.101) { 90};
\draw[gp path] (1.504,7.866)--(1.684,7.866);
\draw[gp path] (6.847,7.866)--(6.667,7.866);
\node[gp node right] at (1.320,7.866) { 95};
\draw[gp path] (1.504,8.630)--(1.684,8.630);
\draw[gp path] (6.847,8.630)--(6.667,8.630);
\node[gp node right] at (1.320,8.630) { 100};
\draw[gp path] (1.504,0.985)--(1.504,1.165);
\draw[gp path] (1.504,8.630)--(1.504,8.450);
\node[gp node center] at (1.504,0.677) { 0};
\draw[gp path] (2.573,0.985)--(2.573,1.165);
\draw[gp path] (2.573,8.630)--(2.573,8.450);
\node[gp node center] at (2.573,0.677) { 5};
\draw[gp path] (3.641,0.985)--(3.641,1.165);
\draw[gp path] (3.641,8.630)--(3.641,8.450);
\node[gp node center] at (3.641,0.677) { 10};
\draw[gp path] (4.710,0.985)--(4.710,1.165);
\draw[gp path] (4.710,8.630)--(4.710,8.450);
\node[gp node center] at (4.710,0.677) { 15};
\draw[gp path] (5.778,0.985)--(5.778,1.165);
\draw[gp path] (5.778,8.630)--(5.778,8.450);
\node[gp node center] at (5.778,0.677) { 20};
\draw[gp path] (6.847,0.985)--(6.847,1.165);
\draw[gp path] (6.847,8.630)--(6.847,8.450);
\node[gp node center] at (6.847,0.677) { 25};
\draw[gp path] (1.504,8.630)--(1.504,0.985)--(6.847,0.985)--(6.847,8.630)--cycle;
\node[gp node center,rotate=-270] at (0.246,4.807) {Perplexity};
\node[gp node center] at (4.175,0.215) {Steps (in millions examples)};
\node[gp node right,font={\fontsize{8pt}{9.6pt}\selectfont}] at (5.379,8.296) {Normal std=10};
\gpcolor{rgb color={1.000,0.647,0.000}}
\gpsetlinetype{gp lt plot 0}
\gpsetlinewidth{2.00}
\draw[gp path] (5.563,8.296)--(6.479,8.296);
\draw[gp path] (2.119,8.630)--(2.145,8.441)--(2.209,8.492)--(2.273,8.262)--(2.337,8.497)%
  --(2.402,7.708)--(2.466,7.837)--(2.530,7.508)--(2.594,7.531)--(2.658,7.059)--(2.722,7.468)%
  --(2.786,7.243)--(2.850,7.320)--(2.915,7.087)--(2.979,7.019)--(3.043,6.841)--(3.107,6.856)%
  --(3.171,7.093)--(3.235,6.887)--(3.299,7.034)--(3.363,6.717)--(3.427,6.452)--(3.492,6.716)%
  --(3.556,6.572)--(3.620,6.490)--(3.684,6.432)--(3.748,6.453)--(3.812,6.337)--(3.876,6.522)%
  --(3.940,6.301)--(4.005,6.035)--(4.069,6.046)--(4.133,5.964)--(4.197,6.191)--(4.261,6.014)%
  --(4.325,6.030)--(4.389,5.979)--(4.453,5.862)--(4.517,5.989)--(4.582,5.848)--(4.646,5.719)%
  --(4.710,5.844)--(4.774,5.947)--(4.838,5.936)--(4.902,5.768)--(4.966,5.763)--(5.030,5.723)%
  --(5.094,5.697)--(5.159,5.745)--(5.223,5.735)--(5.287,5.504)--(5.351,5.688)--(5.415,5.537)%
  --(5.479,5.620)--(5.543,5.559)--(5.607,5.617)--(5.672,5.392)--(5.736,5.485)--(5.800,5.337)%
  --(5.864,5.385)--(5.928,5.369)--(5.992,5.387)--(6.056,5.346)--(6.120,5.447)--(6.184,5.296)%
  --(6.249,5.153)--(6.313,5.381)--(6.377,5.207)--(6.441,5.125)--(6.505,5.300)--(6.569,5.372)%
  --(6.633,5.052)--(6.697,5.053)--(6.761,5.067)--(6.826,4.980);
\gpcolor{color=gp lt color border}
\node[gp node right,font={\fontsize{8pt}{9.6pt}\selectfont}] at (5.379,7.988) {Normal std=1};
\gpcolor{rgb color={1.000,0.000,0.000}}
\gpsetlinetype{gp lt plot 1}
\draw[gp path] (5.563,7.988)--(6.479,7.988);
\draw[gp path] (1.676,8.630)--(1.681,8.552)--(1.769,7.588)--(1.856,7.066)--(1.945,6.797)%
  --(2.034,6.512)--(2.121,6.412)--(2.210,6.040)--(2.297,5.994)--(2.386,5.572)--(2.474,5.311)%
  --(2.562,5.303)--(2.651,5.308)--(2.739,5.120)--(2.827,4.876)--(2.916,4.935)--(3.004,4.819)%
  --(3.094,4.641)--(3.183,4.683)--(3.271,4.625)--(3.360,4.275)--(3.448,4.279)--(3.537,4.354)%
  --(3.626,4.270)--(3.714,4.169)--(3.803,4.271)--(3.892,4.085)--(3.980,3.911)--(4.070,4.083)%
  --(4.158,3.858)--(4.249,3.811)--(4.339,3.777)--(4.428,3.945)--(4.518,3.656)--(4.606,3.801)%
  --(4.697,3.558)--(4.787,3.821)--(4.875,3.785)--(4.967,3.714)--(5.058,3.585)--(5.149,3.442)%
  --(5.242,3.621)--(5.332,3.543)--(5.424,3.504)--(5.516,3.458)--(5.607,3.239)--(5.699,3.464)%
  --(5.790,3.239)--(5.883,3.255)--(5.975,3.333)--(6.066,3.301)--(6.157,3.052)--(6.247,3.229)%
  --(6.338,3.334)--(6.429,3.051)--(6.518,3.134)--(6.609,3.163)--(6.699,3.235)--(6.787,3.174)%
  --(6.847,3.193);
\gpcolor{color=gp lt color border}
\node[gp node right,font={\fontsize{8pt}{9.6pt}\selectfont}] at (5.379,7.680) {Ones};
\gpcolor{rgb color={0.000,0.000,1.000}}
\gpsetlinetype{gp lt plot 2}
\draw[gp path] (5.563,7.680)--(6.479,7.680);
\draw[gp path] (1.763,8.630)--(1.765,8.610)--(1.852,7.475)--(1.938,6.844)--(2.025,6.954)%
  --(2.112,6.204)--(2.199,5.945)--(2.286,5.615)--(2.373,5.418)--(2.459,5.197)--(2.546,5.527)%
  --(2.633,5.041)--(2.720,4.912)--(2.807,4.651)--(2.894,4.714)--(2.980,4.945)--(3.067,4.455)%
  --(3.154,4.572)--(3.241,4.518)--(3.328,4.449)--(3.414,4.304)--(3.501,4.021)--(3.588,4.040)%
  --(3.675,4.045)--(3.761,3.862)--(3.848,3.900)--(3.935,3.854)--(4.022,3.849)--(4.108,3.768)%
  --(4.195,3.836)--(4.282,3.593)--(4.369,3.570)--(4.455,3.582)--(4.542,3.499)--(4.629,3.697)%
  --(4.716,3.406)--(4.802,3.492)--(4.889,3.220)--(4.976,3.393)--(5.063,3.497)--(5.149,3.259)%
  --(5.236,3.176)--(5.323,2.996)--(5.410,3.349)--(5.496,3.288)--(5.583,3.326)--(5.670,3.206)%
  --(5.757,3.127)--(5.843,2.938)--(5.930,3.055)--(6.017,2.843)--(6.104,2.835)--(6.190,2.941)%
  --(6.277,2.785)--(6.364,2.842)--(6.450,2.926)--(6.537,2.876)--(6.624,2.785)--(6.711,2.843)%
  --(6.797,2.704);
\gpcolor{color=gp lt color border}
\node[gp node right,font={\fontsize{8pt}{9.6pt}\selectfont}] at (5.379,7.372) {Remaining methods};
\gpcolor{rgb color={0.000,1.000,0.000}}
\gpsetlinetype{gp lt plot 3}
\draw[gp path] (5.563,7.372)--(6.479,7.372);
\draw[gp path] (1.568,8.364)--(1.632,6.810)--(1.696,6.126)--(1.760,5.635)--(1.825,5.299)%
  --(1.889,4.827)--(1.953,4.557)--(2.017,4.567)--(2.081,4.243)--(2.145,3.987)--(2.209,4.072)%
  --(2.273,4.024)--(2.337,3.819)--(2.402,3.643)--(2.466,3.667)--(2.530,3.495)--(2.594,3.600)%
  --(2.658,3.200)--(2.722,3.475)--(2.786,3.098)--(2.850,3.320)--(2.915,2.929)--(2.979,3.177)%
  --(3.043,3.078)--(3.107,3.087)--(3.171,2.918)--(3.235,2.813)--(3.299,2.899)--(3.363,2.860)%
  --(3.427,2.494)--(3.492,3.249)--(3.556,2.605)--(3.620,2.722)--(3.684,2.666)--(3.748,2.605)%
  --(3.812,2.537)--(3.876,2.610)--(3.940,2.598)--(4.005,2.515)--(4.069,2.411)--(4.133,2.558)%
  --(4.197,2.457)--(4.261,2.529)--(4.325,2.383)--(4.389,2.316)--(4.453,2.311)--(4.517,2.382)%
  --(4.582,2.143)--(4.646,2.215)--(4.710,2.134)--(4.774,2.256)--(4.838,2.322)--(4.902,2.299)%
  --(4.966,2.146)--(5.030,2.201)--(5.094,2.044)--(5.159,2.128)--(5.223,2.260)--(5.287,2.098)%
  --(5.351,2.183)--(5.415,1.889)--(5.479,2.063)--(5.543,2.106)--(5.607,2.098)--(5.672,2.152)%
  --(5.736,2.083)--(5.800,2.019)--(5.864,1.924)--(5.928,1.889)--(5.992,1.997)--(6.056,1.932)%
  --(6.120,2.046)--(6.184,2.085)--(6.249,1.850)--(6.313,1.977)--(6.377,1.958)--(6.441,2.044)%
  --(6.505,1.965)--(6.569,1.956)--(6.633,2.014)--(6.697,1.696)--(6.761,1.856)--(6.826,1.729);
\gpsetlinetype{gp lt plot 4}
\draw[gp path] (1.591,7.708)--(1.678,6.099)--(1.765,5.113)--(1.852,4.754)--(1.939,4.398)%
  --(2.025,4.422)--(2.111,4.158)--(2.198,3.988)--(2.284,3.754)--(2.371,3.759)--(2.457,3.777)%
  --(2.544,3.476)--(2.630,3.671)--(2.717,3.222)--(2.803,3.124)--(2.890,3.143)--(2.976,3.065)%
  --(3.062,3.013)--(3.149,2.782)--(3.235,3.019)--(3.322,2.839)--(3.408,2.789)--(3.495,2.713)%
  --(3.581,2.666)--(3.667,2.755)--(3.754,2.664)--(3.840,2.639)--(3.927,2.482)--(4.013,2.422)%
  --(4.100,2.500)--(4.186,2.598)--(4.272,2.319)--(4.359,2.401)--(4.445,2.387)--(4.532,2.384)%
  --(4.618,2.361)--(4.704,2.323)--(4.791,2.292)--(4.877,2.355)--(4.964,2.252)--(5.050,2.156)%
  --(5.137,2.101)--(5.223,2.111)--(5.309,2.033)--(5.396,2.018)--(5.482,2.308)--(5.569,2.022)%
  --(5.655,2.014)--(5.742,1.861)--(5.828,2.092)--(5.914,2.014)--(6.001,1.898)--(6.087,2.028)%
  --(6.174,1.877)--(6.260,1.813)--(6.347,2.092)--(6.433,1.877)--(6.519,2.012)--(6.606,1.893)%
  --(6.692,2.110)--(6.779,1.898)--(6.847,1.884);
\gpsetlinetype{gp lt plot 5}
\draw[gp path] (1.578,8.630)--(1.632,7.266)--(1.696,6.547)--(1.760,5.924)--(1.825,5.633)%
  --(1.889,5.099)--(1.953,4.735)--(2.017,4.830)--(2.081,4.475)--(2.145,4.212)--(2.209,4.273)%
  --(2.273,4.260)--(2.337,4.026)--(2.402,3.807)--(2.466,3.865)--(2.530,3.704)--(2.594,3.749)%
  --(2.658,3.286)--(2.722,3.674)--(2.786,3.233)--(2.850,3.467)--(2.915,3.124)--(2.979,3.456)%
  --(3.043,3.342)--(3.107,3.231)--(3.171,3.108)--(3.235,2.984)--(3.299,3.081)--(3.363,3.057)%
  --(3.427,2.655)--(3.492,3.317)--(3.556,2.790)--(3.620,2.877)--(3.684,2.884)--(3.748,2.746)%
  --(3.812,2.670)--(3.876,2.802)--(3.940,2.644)--(4.005,2.661)--(4.069,2.636)--(4.133,2.782)%
  --(4.197,2.645)--(4.261,2.694)--(4.325,2.485)--(4.389,2.519)--(4.453,2.402)--(4.517,2.476)%
  --(4.582,2.320)--(4.646,2.437)--(4.710,2.358)--(4.774,2.405)--(4.838,2.528)--(4.902,2.426)%
  --(4.966,2.287)--(5.030,2.314)--(5.094,2.095)--(5.159,2.314)--(5.223,2.390)--(5.287,2.228)%
  --(5.351,2.293)--(5.415,2.049)--(5.479,2.242)--(5.543,2.230)--(5.607,2.141)--(5.672,2.251)%
  --(5.736,2.201)--(5.800,2.071)--(5.864,2.064)--(5.928,1.948)--(5.992,2.112)--(6.056,2.099)%
  --(6.120,2.137)--(6.184,2.165)--(6.249,1.889)--(6.313,2.025)--(6.377,2.057)--(6.441,2.051)%
  --(6.505,2.032)--(6.569,2.071)--(6.633,2.152)--(6.697,1.802)--(6.761,2.077)--(6.826,1.875);
\gpsetlinetype{gp lt plot 6}
\draw[gp path] (1.568,8.348)--(1.632,6.779)--(1.696,6.064)--(1.760,5.623)--(1.825,5.245)%
  --(1.889,4.851)--(1.953,4.521)--(2.017,4.555)--(2.081,4.305)--(2.145,4.040)--(2.209,4.056)%
  --(2.273,4.079)--(2.337,3.810)--(2.402,3.670)--(2.466,3.734)--(2.530,3.451)--(2.594,3.553)%
  --(2.658,3.206)--(2.722,3.419)--(2.786,3.140)--(2.850,3.223)--(2.915,2.927)--(2.979,3.165)%
  --(3.043,3.051)--(3.107,3.061)--(3.171,2.944)--(3.235,2.848)--(3.299,2.835)--(3.363,2.915)%
  --(3.427,2.521)--(3.492,3.217)--(3.556,2.701)--(3.620,2.699)--(3.684,2.783)--(3.748,2.611)%
  --(3.812,2.611)--(3.876,2.572)--(3.940,2.601)--(4.005,2.429)--(4.069,2.442)--(4.133,2.551)%
  --(4.197,2.507);
\gpsetlinetype{gp lt plot 7}
\draw[gp path] (1.568,8.239)--(1.632,6.697)--(1.696,6.056)--(1.760,5.560)--(1.825,5.221)%
  --(1.889,4.850)--(1.953,4.526)--(2.017,4.525)--(2.081,4.402)--(2.145,4.018)--(2.209,4.088)%
  --(2.273,4.050)--(2.337,3.738)--(2.402,3.656)--(2.466,3.668)--(2.530,3.453)--(2.594,3.578)%
  --(2.658,3.152)--(2.722,3.489)--(2.786,3.148)--(2.850,3.268)--(2.915,2.911)--(2.979,3.227)%
  --(3.043,3.119)--(3.107,3.037)--(3.171,3.054)--(3.235,2.874)--(3.299,2.926)--(3.363,2.834)%
  --(3.427,2.559)--(3.492,3.184)--(3.556,2.654)--(3.620,2.664)--(3.684,2.680)--(3.748,2.677)%
  --(3.812,2.600)--(3.876,2.599)--(3.940,2.640)--(4.005,2.412)--(4.069,2.497)--(4.133,2.610)%
  --(4.197,2.522)--(4.261,2.532)--(4.325,2.398)--(4.389,2.296)--(4.453,2.323)--(4.517,2.430)%
  --(4.582,2.282)--(4.646,2.344)--(4.710,2.219)--(4.774,2.358)--(4.838,2.364)--(4.902,2.386)%
  --(4.966,2.133)--(5.030,2.341)--(5.094,2.053)--(5.159,2.258)--(5.223,2.321)--(5.287,2.099)%
  --(5.351,2.193)--(5.415,2.109)--(5.479,2.212)--(5.543,2.175)--(5.607,2.224)--(5.672,2.103)%
  --(5.736,2.074)--(5.800,1.989)--(5.864,1.944)--(5.928,1.990)--(5.992,2.088)--(6.056,2.057)%
  --(6.120,2.072)--(6.184,2.124)--(6.249,1.860)--(6.313,1.899)--(6.377,1.911)--(6.441,1.968)%
  --(6.505,2.006)--(6.569,2.012)--(6.633,2.009)--(6.697,1.690)--(6.761,1.956)--(6.826,1.807);
\gpsetlinetype{gp lt plot 0}
\draw[gp path] (1.593,8.258)--(1.684,6.245)--(1.773,5.451)--(1.863,5.032)--(1.954,4.669)%
  --(2.044,4.285)--(2.135,4.164)--(2.225,3.935)--(2.315,3.805)--(2.405,3.667)--(2.496,3.568)%
  --(2.586,3.462)--(2.676,3.289)--(2.767,3.182)--(2.857,3.023)--(2.947,3.071)--(3.038,3.074)%
  --(3.128,3.012)--(3.219,3.345)--(3.308,2.761)--(3.399,2.834)--(3.489,2.707)--(3.580,2.694)%
  --(3.672,2.550)--(3.763,2.582)--(3.855,2.520)--(3.946,2.538)--(4.038,2.489)--(4.129,2.613)%
  --(4.221,2.674)--(4.313,2.318)--(4.405,2.369)--(4.497,2.256)--(4.588,2.214)--(4.680,2.116)%
  --(4.772,2.299)--(4.863,2.180)--(4.955,2.157)--(5.045,2.030)--(5.137,2.065)--(5.227,2.133)%
  --(5.319,2.166)--(5.411,1.925)--(5.502,2.149)--(5.594,2.198)--(5.685,2.071)--(5.777,2.038)%
  --(5.868,1.998)--(5.960,1.893)--(6.052,1.924)--(6.143,2.000)--(6.236,1.837)--(6.327,1.779)%
  --(6.419,1.733)--(6.510,1.866)--(6.601,1.961)--(6.691,1.890)--(6.779,1.724);
\gpcolor{color=gp lt color border}
\node[gp node right,font={\fontsize{8pt}{9.6pt}\selectfont}] at (5.379,7.064) {Word2Vec};
\gpcolor{rgb color={0.647,0.165,0.165}}
\gpsetlinetype{gp lt plot 1}
\draw[gp path] (5.563,7.064)--(6.479,7.064);
\draw[gp path] (1.592,6.977)--(1.680,5.607)--(1.767,5.001)--(1.854,4.534)--(1.942,4.356)%
  --(2.029,3.898)--(2.116,3.654)--(2.203,3.506)--(2.291,3.306)--(2.378,3.183)--(2.465,3.006)%
  --(2.553,3.057)--(2.640,2.960)--(2.727,2.954)--(2.815,2.956)--(2.903,2.849)--(2.991,2.633)%
  --(3.078,2.827)--(3.166,2.558)--(3.254,2.383)--(3.344,2.443)--(3.435,2.425)--(3.526,2.273)%
  --(3.618,2.264)--(3.709,2.506)--(3.801,2.538)--(3.892,2.202)--(3.985,2.064)--(4.077,2.210)%
  --(4.168,2.114)--(4.260,2.149)--(4.352,2.150)--(4.444,2.137)--(4.535,2.151)--(4.626,2.058)%
  --(4.718,1.863)--(4.808,1.900)--(4.900,1.913)--(4.990,1.786)--(5.081,2.072)--(5.172,2.065)%
  --(5.263,1.966)--(5.355,1.833)--(5.446,1.839)--(5.538,1.733)--(5.629,1.739)--(5.720,1.968)%
  --(5.812,1.658)--(5.903,1.784)--(5.993,1.668)--(6.083,1.643)--(6.175,1.519)--(6.266,1.730)%
  --(6.356,1.583)--(6.448,1.740)--(6.537,1.597)--(6.629,1.613)--(6.719,1.595)--(6.810,1.763);
\gpcolor{rgb color={0.000,1.000,0.000}}
\gpsetlinetype{gp lt plot 2}
\draw[gp path] (1.591,7.348)--(1.678,6.141)--(1.765,5.426)--(1.851,5.056)--(1.938,4.587)%
  --(2.025,4.512)--(2.112,4.319)--(2.198,4.194)--(2.285,3.930)--(2.372,4.010)--(2.458,3.788)%
  --(2.545,3.528)--(2.632,3.674)--(2.718,3.470)--(2.805,3.606)--(2.892,3.094)--(2.978,2.969)%
  --(3.065,3.130)--(3.152,3.027)--(3.238,3.167)--(3.325,2.888)--(3.412,2.709)--(3.498,2.825)%
  --(3.585,2.653)--(3.672,2.623)--(3.758,2.631)--(3.845,2.497)--(3.931,2.640)--(4.018,2.627)%
  --(4.105,2.489)--(4.191,2.526)--(4.278,2.297)--(4.365,2.514)--(4.451,2.443)--(4.538,2.277)%
  --(4.624,2.396)--(4.711,2.268)--(4.798,2.261)--(4.884,2.448)--(4.971,2.304)--(5.057,2.349)%
  --(5.144,2.297)--(5.231,2.235)--(5.317,2.330)--(5.404,2.186)--(5.491,2.252)--(5.577,2.270)%
  --(5.664,1.974)--(5.750,2.066)--(5.837,2.225)--(5.923,2.064)--(6.010,1.950)--(6.097,1.961)%
  --(6.183,1.953)--(6.270,2.000)--(6.357,1.850)--(6.444,1.890)--(6.531,1.860)--(6.618,2.039)%
  --(6.705,1.874)--(6.792,2.111);
\gpcolor{color=gp lt color border}
\node[gp node right,font={\fontsize{8pt}{9.6pt}\selectfont}] at (5.379,6.756) {Selftrain};
\gpcolor{rgb color={0.753,0.502,1.000}}
\gpsetlinetype{gp lt plot 3}
\draw[gp path] (5.563,6.756)--(6.479,6.756);
\draw[gp path] (1.568,6.414)--(1.632,5.396)--(1.696,4.810)--(1.760,4.500)--(1.825,4.208)%
  --(1.889,3.820)--(1.953,3.604)--(2.017,3.572)--(2.081,3.408)--(2.145,3.176)--(2.209,3.263)%
  --(2.273,3.283)--(2.337,2.940)--(2.402,2.872)--(2.466,2.796)--(2.530,2.715)--(2.594,2.727)%
  --(2.658,2.467)--(2.722,2.684)--(2.786,2.355)--(2.850,2.689)--(2.915,2.335)--(2.979,2.555)%
  --(3.043,2.483)--(3.107,2.482)--(3.171,2.463)--(3.235,2.240)--(3.299,2.396)--(3.363,2.289)%
  --(3.427,1.953)--(3.492,2.609)--(3.556,2.128)--(3.620,2.265)--(3.684,2.214)--(3.748,2.132)%
  --(3.812,2.052)--(3.876,2.143)--(3.940,2.178)--(4.005,2.071)--(4.069,2.116)--(4.133,2.167)%
  --(4.197,2.079)--(4.261,2.193)--(4.325,1.881)--(4.389,1.879)--(4.453,1.849)--(4.517,1.976)%
  --(4.582,1.825)--(4.646,1.867)--(4.710,1.843)--(4.774,1.892)--(4.838,1.985)--(4.902,1.918)%
  --(4.966,1.743)--(5.030,1.843)--(5.094,1.712)--(5.159,1.827)--(5.223,1.973)--(5.287,1.820)%
  --(5.351,1.782)--(5.415,1.626)--(5.479,1.780)--(5.543,1.861)--(5.607,1.791)--(5.672,1.798)%
  --(5.736,1.806)--(5.800,1.600)--(5.864,1.611)--(5.928,1.607)--(5.992,1.686)--(6.056,1.734)%
  --(6.120,1.779)--(6.184,1.780)--(6.249,1.569)--(6.313,1.645)--(6.377,1.738)--(6.441,1.773)%
  --(6.505,1.802)--(6.569,1.752)--(6.633,1.738)--(6.697,1.555)--(6.761,1.727)--(6.826,1.641);
\gpcolor{color=gp lt color border}
\gpsetlinetype{gp lt border}
\gpsetlinewidth{1.00}
\draw[gp path] (1.504,8.630)--(1.504,0.985)--(6.847,0.985)--(6.847,8.630)--cycle;
\gpdefrectangularnode{gp plot 1}{\pgfpoint{1.504cm}{0.985cm}}{\pgfpoint{6.847cm}{8.630cm}}
\end{tikzpicture}